\documentclass{article}

\usepackage{arxiv}

\usepackage[utf8]{inputenc} 
\usepackage[T1]{fontenc}    
\usepackage{hyperref}       
\hypersetup{
    colorlinks = true,    
    linkcolor = black,    
    citecolor = black,    
    urlcolor = blue       
}
\usepackage{url}            
\usepackage{booktabs}       
\usepackage{amsfonts}       
\usepackage{nicefrac}       
\usepackage{microtype}      
\usepackage{lipsum}		
\usepackage{graphicx}
\usepackage{natbib}
\usepackage{doi}
\usepackage{booktabs}
\usepackage{multirow}
\usepackage[table]{xcolor}
\usepackage{array}
\renewcommand{\arraystretch}{1.2}

\newcolumntype{L}[1]{>{\raggedright\arraybackslash}p{#1}}
\newcolumntype{C}[1]{>{\centering\arraybackslash}p{#1}}

\newcommand{\NA}{\cellcolor{gray!20}\rule{0pt}{2.6ex}\textsc{N/A}}

\title{Metaphor identification using large language models: A comparison of rag, prompt engineering, and fine-tuning}

\author{
Matteo Fuoli \\
Department of Linguistics and Communication\\
University of Birmingham\\
\texttt{m.fuoli@bham.ac.uk}
\And
Weihang Huang \\
Department of Linguistics and Communication\\
University of Birmingham\\
\texttt{wxh207@student.bham.ac.uk}
\And
Jeannette Littlemore \\
Department of Linguistics and Communication\\
University of Birmingham\\
\texttt{j.m.littlemore@bham.ac.uk}
\AND
Sarah Turner \\
Centre for Arts, Memory \& Communities\\
Coventry University\\
\texttt{Sarah.Turner@coventry.ac.uk}
\And
Ellen Wilding \\
Department of Linguistics and Communication\\
University of Birmingham\\
\texttt{esw217@student.bham.ac.uk}
}

\renewcommand{\shorttitle}{Metaphor identification using large language models}

\hypersetup{
pdftitle={A template for the arxiv style},
pdfsubject={q-bio.NC, q-bio.QM},
pdfauthor={David S.~Hippocampus, Elias D.~Striatum},
pdfkeywords={First keyword, Second keyword, More},
}

\begin{document}
\maketitle

\begin{abstract}
Metaphor is a pervasive feature of discourse and a powerful lens for examining cognition, emotion, and ideology. Large-scale analysis, however, has been constrained by the need for manual annotation due to the context-sensitive nature of metaphor. This study investigates the potential of large language models (LLMs) to automate metaphor identification in full texts. We compare three methods: (i) retrieval-augmented generation (RAG), where the model is provided with a codebook and instructed to annotate texts based on its rules and examples; (ii) prompt engineering, where we design task-specific verbal instructions; and (iii) fine-tuning, where the model is trained on hand-coded texts to optimize performance. Within prompt engineering, we test zero-shot, few-shot, and chain-of-thought strategies. Our results show that state-of-the-art closed-source LLMs can achieve high accuracy, with fine-tuning yielding a median F1 score of 0.79. A comparison of human and LLM outputs reveals that most discrepancies are systematic, reflecting well-known grey areas and conceptual challenges in metaphor theory. We propose that LLMs can be used to at least partly automate metaphor identification and can serve as a testbed for developing and refining metaphor identification protocols and the theory that underpins them.
\end{abstract}


\section{Introduction}
\label{sec:intro}
A large body of research has shown metaphor to be a pervasive feature of discourse and a fundamental tool for communication and interaction. By drawing on more concrete, familiar, easily understood ideas, speakers can articulate their subjective experiences in ways that are more accessible to others. Metaphor analysis thus provides a valuable lens for exploring important research questions with profound individual and social implications, such as patients' experiences of cancer \citep{sdhpr17} or public perceptions of immigration and national identity \citep{m16}. However, metaphor analysis is an extremely time-consuming process because there are currently no reliable methods for automating the crucial yet challenging task of identifying metaphorical expressions in text. The need for manual text annotation, a labour-intensive and error-prone process, inherently limits the scalability of metaphor analysis and, by extension, the degree to which findings can be generalized.

Recent studies in natural language processing (NLP) have shown that large language models (LLMs) – advanced AI systems trained on massive text corpora – can be used to identify metaphors automatically with promising levels of accuracy \citep{prr24, txm24, ych24, hicke2024science}. However, these studies define the task narrowly, focusing on classifying single, often pre-selected words as either literal or metaphorical within decontextualized sentences. This narrow operationalization not only yields outputs of limited value to linguists, but also raises concerns about validity, as metaphors often go beyond single words and can only be properly interpreted by considering the surrounding text. Moreover, existing research explores only a limited set of strategies for deploying LLMs in this task, leaving much of the methodological landscape unexplored.

In this study, we assess the performance of LLMs in the task of identifying and annotating metaphor in full texts. We conduct a series of experiments to compare three distinct methods: (i) `retrieval-augmented generation' (RAG), where we provide the model with our codebook as a knowledge resource and instruct it to annotate our corpus based on the rules, principles, and examples included in it; (ii) `prompt engineering', where we design verbal instructions for the model to perform the task; and (iii) `fine-tuning', where we train the model on a sample of hand-coded texts to optimize its performance. Within the prompt engineering strand, we experiment with three common prompting strategies: `zero-shot', `few-shot', and `chain-of-thought'. We also examine the effects of varying the number and type of examples included in the prompt to guide the model in the identification task. We conduct these experiments using 10 of the most advanced closed- and open-source LLMs available at the time of writing. We apply these LLMs to a corpus of film reviews hand-coded using a `phraseological' approach that captures both single- and multi-word metaphorical expressions \citep{flt22}.

This study advances both metaphor studies and NLP by providing the first evaluation of LLMs for full-text metaphor identification and annotation, alongside a comparison of three core methodological approaches. If LLMs can perform this task accurately, they could transform metaphor research by providing an accessible tool for efficiently annotating large text corpora, enabling researchers to scale up and make their analyses more generalizable, and redirect resources from low-level annotation tasks to higher-level interpretation and theory-building. LLMs offer a key advantage over other computational methods: they can be guided using natural language prompts, thus empowering researchers without advanced programming expertise to automate one of the most tedious and time-consuming parts of their work. In addition to this methodological contribution, our study offers new theoretical insights into metaphor by analysing the LLM annotations and comparing them with those of human coders. We show that this process can help reveal areas where current metaphor identification protocols – and the theory that underpins them – may be refined.

\section{Background}
\label{sec:background}
This section provides the essential theoretical and methodological background for this study. We begin by defining the analytical task of metaphor identification and reviewing existing approaches, before outlining our own. Next, we summarize previous work on LLM-assisted metaphor identification.

\subsection{Identifying metaphors in discourse}
\label{sec:identifying}
Metaphor can be defined in different ways according to whether one is interested in approaching it at the linguistic level, at the conceptual level, or at the level of discourse \citep{s09}. However, in practice, it is impossible to perceive a clear separation between these different levels. In this study, therefore, we are interested in providing an automated identification procedure that can be used regardless of the level. In order to do this, we need to start with a definition of metaphor that works on all three levels. We therefore define metaphor as a tool for reasoning, or for communication, that involves comparing one entity (the `topic') to another entity (the `vehicle'), in order to express an opinion about that entity, to communicate one's feelings about it, and/or to render it more easily understood. We acknowledge the presence of underlying systematic \citep{c10} and/or conceptual \citep{lj80} mappings that underpin a good number of surface-level ``linguistic'' metaphors, although this is not a focus of the current study. 

A long-standing problem faced by metaphor researchers has been the need to establish a rigorous, efficient, and replicable procedure to identify metaphors in discourse. Various solutions have been proposed to this problem, which has led to the development of a number of different approaches. These approaches have advantages and disadvantages. Approaches such as the MIP \citep{g07} or the MIPVU procedure \citep{sdkkh10} involve taking every lexical unit in the text and establishing whether it is being used metaphorically. This involves checking to see whether its contextual meaning can be understood through comparison with a more basic entity. This is a rigorous approach involving the use of dictionaries to establish the different senses of the lexical unit, and as such its results have been shown to be highly replicable. The problem with this approach is that it is very time-consuming, and it is therefore difficult to apply to large corpora. Furthermore, identifying meaningful metaphorical items often requires an additional level of coding, which considers how individual lexical items flagged as potentially metaphorical combine to form a single metaphorical meaning unit within a phrase or longer stretch of text. 

Other researchers have adapted their metaphor identification methodology based on their research questions, which has led to a greater or lesser adherence to the MIP(VU). For example, studies such as \citet{lt20} have adopted a more phrase-based approach to metaphor, reminiscent of Cameron's \citeyear{c03} `Metaphor Identification through Vehicle Terms' (MIV). The MIV procedure consists of finding pieces of language in the text that have the potential to be interpreted metaphorically. Cameron defines `vehicle term' as a word or phrase that ``stands out clearly as incongruous or anomalous in its discourse context'' \citep[p. 4]{c10-1}. For Cameron, there are two necessary conditions for the identification of linguistic metaphor: a contrast in meaning between the vehicle and topic and a transfer or connection of meaning between them \citep[pp. 59-60]{c03}. A similar approach to metaphor identification has also been taken by \citet{jo23} in their Procedure for Identifying Metaphorical Scenes (PIMS). Used specifically for the identification of metaphor in prepositional constructions, this method allows the analyst to identify metaphorical meaning that extends over phrases or longer stretches of text. It focuses on the scenes evoked by these phrases and allows the analyst to establish whether the scenes are metaphorical, non-metaphorical, or ambiguous.

The flexibility of this latter set of approaches allows them to be tailored to the characteristics of the data, and to focus on the phraseological meaning. These methods bring to the fore the ways in which metaphor functions in context across broader stretches of text. However, this flexibility may engender issues with replicability, as it is not always easy to see where a metaphor begins and ends. Also, when the metaphor identification procedure has been developed iteratively in order to meet the needs of the research questions, it may be difficult for a third party to apply the procedure in the same way.

To illustrate the differences between approaches that focus primarily on individual lexical units (which are usually words) and approaches that focus more on metaphor at the level of the phrase, let us consider the following example from \citet{tl23} which is taken from an interview with a bereaved parent talking about how she felt when her child died:

\begin{quote}
We were just forcibly thrown into this awful universe of just misery and pointlessness.
\end{quote}

If an analyst were to apply a procedure such as the MIP(VU), they would only identify lexical units such as \emph{thrown}, \emph{universe}, and possibly \emph{forcibly} as metaphorical. While this analysis may be useful if one were interested in measuring metaphoric density, it does not provide insight into the holistic experience that the parent is describing. A more scene-based approach to metaphor analysis would take the whole utterance as a single metaphorical instance. This provides richer insight into the nature of the experience that the parent is describing but does allow for a certain degree of subjectivity in deciding where the metaphor begins and ends. 

For the reasons outlined above, the approach to metaphor identification that we take in this study aligns more with this latter set of approaches. The procedure employed is described in detail in \citet{flt22}. We began by reading the entire text to establish a general understanding of the meaning. We then identified meaning units at the level of phrase following Cameron's \citeyearpar{c03} VIP. For each meaning unit, we established its meaning in context (i.e. its contextual meaning, taking into account what comes before and after the meaning unit). Having done so, we determined whether the phrase had a more basic contemporary meaning in other contexts than the one in the given context. For our purposes, basic meanings tend to be more concrete. If the meaning unit had a more concrete contemporary meaning in other contexts than the given context, and if this contextual meaning could be understood in comparison with it, it was marked as metaphorical. In some cases, metaphors were identified at the level of the single word. However, they often extended beyond single words. This could occur when: 

\begin{enumerate}
    \item The expression was a conventional idiom, e.g. \emph{have your cake and eat it}.
    \item There were hyphenated words which formed a single lexical unit, e.g. \emph{tough-as-nails}.
    \item There was an adjectival entailment of a metaphorically used noun (or an adverbial entailment of a metaphorically used verb) that was internally semantically coherent with the literal sense of the noun or verb. Phrases that were internally coherent were marked as a single metaphor, even when there was a non-metaphorical stretch of texts separating them. However, if there were two distinct ideas in the same metaphorical phrase, these were marked as separate metaphors. E.g. \emph{It's pretty much a \textbf{sunken ship} of a movie} (Ships can sink in the literal world and \emph{sunken} is serving as a premodifier of \emph{ship} in this sentence).
    \item There was an adjectival entailment of a metaphorically used noun (or an adverbial entailment of a metaphorically used verb) that was internally semantically coherent with the metaphorical sense of the noun or verb, but which would not occur in literal language, e.g. the \emph{negative baggage} that the reviewers were likely to have. (\emph{baggage} in its metaphorical sense here can be positive or negative but literal baggage is not conventionally described in these terms).
\end{enumerate}

Phrases were coded as metaphor even when they were signalled with tuning devices such as \emph{like} or \emph{as}. Individual words were not broken down into their metaphorical components. We followed an overarching principle where we kept the length of the annotated text spans to a minimum. 

Despite the presence of these relatively rigorous, replicable procedures for metaphor identification, and despite the relatively high inter-rater reliability scores obtained for some of the procedures, including the approach that we use in the study, it should be acknowledged that a degree of subjectivity is always inherent in metaphor identification. There will always be grey areas about what constitutes a metaphor or not, and what may be experienced as metaphorical by one person may be experienced as literal by another. This makes the practice of inter-rater reliability testing all the more important, as it is through comparing annotations made independently by different analysts that we can assess the transparency and robustness of our annotation criteria and gain useful insights for refining them. However, performing such tests is not always feasible due to the significant time investment required from multiple people. To address this challenge, we propose that LLMs could be leveraged as artificial reliability coders, provided they are sufficiently accurate in identifying metaphor.

Another key challenge LLMs can help us address is scale. The time-consuming nature of metaphor identification can limit the scope of projects, impeding or even preventing the analysis of large datasets. For these reasons, if it were possible to automate the process of metaphor identification through the use of LLMs, this would make it easier to expand the scope of metaphor studies, allowing metaphor scholars to make more generalisable claims based on their findings.

\subsection{Metaphor identification using LLMs}
\label{sec:LLM-ident}
Automating metaphor identification has long been a central goal in NLP. Traditionally, researchers approached this task – commonly referred to in the field as `metaphor detection' – by developing supervised machine-learning algorithms that label words as literal or metaphorical using linguistic features such as `word embeddings' (numerical representations of word meaning derived from co-occurrence patterns in large text corpora), part-of-speech tags, and word concreteness ratings \citep{gmc23}. The performance of these systems is generally evaluated against hand-coded datasets such as the VU Amsterdam Metaphor Corpus \citep{sdkkh10} or the MOH-X corpus \citep{mst16}, which provide annotations of metaphorical language at the level of individual words. These traditional NLP approaches have achieved solid results, with the best models surpassing 80\% accuracy (for a review and comparison of model performances, see Ge et al., \citeyear{gmc23}).

In recent years, LLMs have emerged as a transformative technology, reshaping research practices and priorities in NLP and driving a paradigm shift across the field. Among the many applications LLMs have been deployed for, metaphor identification has attracted growing interest. Most studies in this area test LLMs on the task of classifying pre-selected single words in sentences as either literal or metaphorical. For example, \citet{prr24} evaluate the performance of OpenAI's LLMs GPT-4o and GPT-4 Turbo on a corpus of Spanish sentences. They compare two variants of the prompt used to instruct the LLMs: one providing only task instructions, and another supplementing these instructions with a brief explainer of metaphor based on Conceptual Metaphor Theory (CMT). They report high average accuracy scores, peaking at 88.29\%. Interestingly, neither the use of the more advanced GPT-4o model nor the inclusion of theory-based instructions led to performance improvements.

Instead of relying on a single prompt, \citet{txm24} introduce an interactive prompting strategy informed by CMT and the MIP procedure designed to help LLMs make more accurate metaphorical judgments by scaffolding their ``reasoning''. Specifically, the model is guided through a sequence of three questions to identify the source domain implied by the target word, determine its target domain within the sentence, and evaluate whether these domains differ, thereby justifying a metaphorical interpretation. This step-by-step approach substantially improves both the accuracy and interpretability of LLM-based metaphor identification, achieving an F1 score as high as 82.59.

Taking a different approach, \citet{ych24} treat GPT-3.5 Turbo not as a direct metaphor classifier, but as a tool for constructing linguistic resources to support metaphor detection. Specifically, they prompt the LLM to generate verb-specific lists of common literal subject-verb and verb-object collocations, which serve as reference points for metaphor identification. This process is further enhanced by generalizing the subjects and objects into broader topic categories using resources like WordNet, LDA topics, and Oxford dictionaries. A verb is classified as metaphorical when the subject and object in the target sentence fail to match any of the topics associated with its literal collocations. Their method achieves fairly good results, with the best performing configuration producing an F1 score of 70.1.

While the studies above test LLMs on the narrow task of judging whether a pre-selected word in a sentence is metaphorical, \citet{liang2025gpt4} use GPT-4 to identify all metaphorical words in a set of sentences from the VU Amsterdam Metaphor Corpus and a smaller set of online news texts. The task follows the MIPVU procedure and focuses solely on conventional metaphors. The authors test a wide range of prompt designs and experiment with varying the number of examples provided to the model (0, 1, 5, and 10). Across all settings, GPT-4's performance was consistently poor, with the best results achieved in the 1-shot condition (F1 = 30.4), far below the performance of previous-generation transformer models such as RoBERTa (F1 77.9–79.7). The authors attribute the weak results to several interacting factors: the style and complexity of the prompts, the limited context provided by single-sentence inputs (some as short as a single word), the uneven representativeness of the examples, and broader limitations of GPT-4. They conclude that prompt-based approaches with LLMs remain unreliable for this task, though future work exploring optimized prompt design or alternative methods such as fine-tuning and RAG may yield better results.

Along similar lines, \citet{hicke2024science} test GPT-3.5 Turbo, GPT-4, and GPT-4o on the task of labelling words as either metaphorical or literal in sentences drawn from the TroFi dataset \citep{bs06} and from \citet{lj80}. Their prompt is adapted almost verbatim from the original MIP procedure, supplemented by two annotated example sentences. In addition to identifying metaphorical words, the models are asked to provide a brief description of each word's more basic meaning, enabling the researchers to assess how well the LLMs can carry out Step 3 of MIP and to make the outputs more interpretable by exposing the models' ``rationale''. The results show that all models perform above chance. GPT-4o performed best, correctly identifying all metaphors in 74\% of sentences from the Lakoff and Johnson dataset and 65\% of the basic meanings. However, all models tended to over-label words as metaphorical and struggled with function words, multiword expressions, and certain metaphor types.

While, with the exception of \citet{liang2025gpt4}, these studies suggest that LLMs hold promise for accurate metaphor identification, the way this task is implemented in NLP, both in traditional and LLM-based approaches, differs significantly from how metaphor scholars practice it. As seen above, instead of training systems to process full texts and identify all metaphorical expressions contained in them, NLP systems are typically designed to classify individual words – often specific, pre-selected target words – as either literal or metaphorical within decontextualized sentences. This approach runs counter to a core principle in metaphor analysis, namely that metaphor is a context-sensitive phenomenon, shaped by both real-world knowledge and the evolving discursive context of a text \citep{s08}. The importance of context is reflected in the very first step of the MIP procedure (often treated as the methodological foundation of NLP approaches), which explicitly instructs annotators to ``read the entire text-discourse to establish a general understanding of the meaning'' \citep[p. 3]{g07}. Another important drawback of current NLP approaches is that they operate strictly at the word level and fail to account for multiword metaphorical expressions, which is problematic for the reasons discussed in Section \ref{sec:identifying}.

The choice to operationalize the task of metaphor detection as the labelling of individual words within individual decontextualized sentences is likely driven by methodological considerations and technical constraints. This setup makes benchmarking of results across studies and iterative improvement easier. But it also reflects the fact that most pre-LLM architectures, within which metaphor detection tasks were originally developed, were constrained by relatively narrow context windows, which made it technically infeasible to process and annotate entire texts at once \citep{dong2023survey}. However, the task, operationalized in this way, has limited construct and ecological validity, and reduces metaphor identification to a narrow engineering challenge rather than a tool that can enhance our understanding of metaphor and facilitate its linguistic analysis across discursive domains.

Our study aims to address these limitations and establish a new framework for assessing the capabilities of LLMs in the task of metaphor identification that better captures the way metaphor actually works in discourse and that produces usable output to support the analysis metaphor across a wide range of contexts. Modern LLMs not only exhibit unprecedented contextual sensitivity but are also able to process texts of considerable length. In the experiments described below, we therefore test LLMs' metaphor identification capabilities on complete texts. The output of the methods we test is an XML tagged corpus, which can serve both as the basis for quantitative analyses of metaphor and as input for a variety of downstream NLP tasks. To overcome the limitations of word-based approaches and enhance the validity of analysis, we adopt a phraseological approach to metaphor identification. As discussed above, this approach is not restricted to individual words but accommodates multiword metaphorical expressions, where warranted by the discursive context.

\section{Methods}
\label{sec:methods}

We evaluated LLMs' ability to identify metaphors by tasking them with labelling all metaphorical expressions in a corpus of IMDb film reviews using \texttt{<Metaphor>} and \texttt{</Metaphor>} XML tags. The corpus, summarized in Table \ref{tab:corpus-details}, was compiled and manually annotated as part of a previous study by two of the authors of this paper based on the criteria discussed in Section \ref{sec:identifying}, achieving robust inter-coder agreement (average Cohen's kappa = 0.85; for details on the annotation and inter-coder agreement procedures, see Fuoli et al., \citeyear{flt22}). We compared three LLM-based methods: RAG, prompt engineering, and fine-tuning. Each method is described in turn below. 

\begin{table}[h!]
\centering
\caption{Corpus details.}
\label{tab:corpus-details}  
\renewcommand{\arraystretch}{1.4} 
\begin{tabular}{l r}
\hline
\textbf{Total number of texts}                    & 94 \\ 
\textbf{Total number of sentences}                & 2,828 \\ 
\textbf{Total number of words}                    & 59,147 \\ 
\textbf{Average text length (in words)}           & 629.4 \\ 
\textbf{Total number of metaphorical expressions} & 2,599 \\ 
\hline
\end{tabular}
\end{table}

RAG allows the LLM to retrieve relevant information from an external resource before generating a response \citep{lpppkgok20}. In this way, the model can ground its output in additional, task-specific knowledge rather than relying solely on the information encoded in its training data. In our implementation, we provided the model with the detailed codebook we developed for annotating the corpus. The codebook, available in the Supplementary Materials, includes our working definition of metaphor, a set of detailed annotation guidelines, and many illustrative examples. We prompted the model to refer to the codebook to decide whether a candidate linguistic expression qualifies as a metaphor. Thus, the RAG-based approach is intended to simulate the normal procedure a human annotator would follow when annotating metaphorical expressions in text.

The prompt used for RAG, shown in the Supplementary Materials, consisted of two components: a `system' prompt and a `user' prompt. Following best-practice guidelines, we used the system prompt to establish the general task framing and define the model's default behaviour and ``persona'' (e.g. OpenAI., \citeyear{o23}). Here, we primed the LLM to act as an expert in metaphor annotation and instructed it to follow the guidelines in the codebook. The last sentence of the system prompt aimed to ensure the LLM returns only the coded text instead of adding extra explanations, comments, or formatting (e.g. ``Here is your tagged text:''). This was done to prevent the need for further text processing once the annotation is complete. The user prompt generally specifies the particular question to be addressed – in this case a request to identify and tag the metaphors in a text.

The second method we evaluated is prompt engineering. Prompt engineering (or ``prompting'') refers to the practice of designing verbal instructions that guide LLMs to produce relevant and accurate outputs. It is one of the most rapidly evolving areas of research in NLP and AI, with a wide range of strategies devised and tested in prior research (for an accessible overview, see Boonstra, \citeyear{b24}). In this study, we adapted three of the most widely used and versatile prompting strategies: zero-shot, few-shot, and chain-of-thought. In zero-shot prompting, the model is provided only with a task description and must rely solely on its built-in knowledge to generate responses \citep{bmrskdoa20}. The prompt we used for this strategy is the same we use for the RAG approach but without the second last sentence in the system prompt instructing the model to consult the codebook.

A common alternative to zero-shot prompting is few-shot prompting, where the model is exposed to a handful of task-related examples along with their expected outputs before attempting to complete the task \citep{bmrskdoa20}. This technique leverages `in-context learning', whereby the model infers patterns from the examples provided and generalizes from them to new, unseen inputs without updating its underlying parameters \citep{bmrskdoa20}. In our case, we provided the model with a series of examples of manually annotated sentences extracted from our corpus. These examples were presented as a sequence of alternating `user' and `assistant' messages: each user message asked the model to identify metaphors in a given sentence, and each assistant message provided the correct annotation. After the sequence of examples, the model received a final user message containing a full text to be annotated. The system prompt was the same as the one used in the zero-shot prompt.

The third prompting strategy we tested is few-shot chain-of-thought prompting (CoT), where the examples provided to the model are supplemented with an explanation of the reasoning behind the correct answer \citep{wwsbxcoz22}. This prompting strategy aims to improve performance by making the reasoning process underlying the task explicit, thus helping the model more effectively internalize and replicate the decision-making steps. Following \citet{hlgjzljydc24}, we presented the model with annotated examples followed by an explanation of the coding based on our annotation protocol. When examples contained multiple metaphorical expressions, each expression was presented and explained in turn. We generated all the explanations accompanying the examples using OpenAI's GPT 4o. We applied a consistent template, shown below, to streamline the process and ensure standardization.

\begin{quote}
The word ``[WORD]'' has a more basic contemporary meaning: in other contexts, it refers to [BASIC MEANING]. In this example, it describes [FIGURATIVE MEANING] by comparing them to something literally [BASIC MEANING], which makes it a metaphorical usage.
\end{quote}

All generated explanations were manually reviewed for accuracy. Only two minor adjustments were made, which highlights GPT 4o's strong inherent capabilities for metaphorical reasoning. 

As in the few-shot prompt, the examples and corresponding explanations were presented as a sequence of alternating `user' and `assistant' messages, followed by a final user message that elicited both the annotations and an explanation of the coding choices. We anticipated that requesting both would improve performance by prompting the model to explicitly ``reason'' about its decisions. In addition, the explanations provide valuable insight into patterns in the LLM's coding and may serve as input for future iterations of this experiment. The system prompt was slightly adapted by rephrasing the final instruction to ensure that the model did not suppress the generation of explanations. 

While available empirical evidence suggests that providing models with examples generally leads to improved performance \citep[e.g.][]{bmrskdoa20, shbchhx24}, it is less clear what the optimal number of examples is. For example, \citet{dqlo23} found that for some tasks, 10-shot prompts did not generate as accurate responses as two-shot prompts. In the context of metaphor identification, \citet{liang2025gpt4} observed that including a single example enhanced LLM performance, but adding further examples yielded no additional gains. To explore how the number of examples affects performance in our task, we varied both the few-shot and CoT prompts by including either four or eight examples.

Another important aspect we considered when selecting the examples was the type of metaphor they included. In addition to identifying metaphorical expressions, \citet{flt22} classify them as either `conventional' or `creative'. Conventional metaphors are those that make use of widely attested comparisons without changing them in any way, whereas creatively used metaphors involve drawing what look like completely new comparisons, or extending existing comparisons in novel ways \citep{tl23}. The ratio of conventional to creative metaphors in the corpus is approximately 9:1. To assess whether LLMs are sensitive to this distinction, we created two versions of the few-shot and CoT prompts, one preserving the original distribution and another balancing the examples to include an equal number of conventional and creative metaphors. 

The complete prompt set is provided in the Supplementary Materials. The wording of all the prompts was iteratively refined through trial runs on small batches of texts and through `meta-prompting', a common technique in which LLMs are asked to evaluate and improve the prompt itself.

The third approach we tested is model fine-tuning. Fine-tuning refers to the process of further training a language model on a task-specific dataset to improve performance \citep{gmslbds20}. Unlike prompt engineering, fine-tuning modifies the model itself by adjusting its internal parameters via supervised learning on labelled data directly relevant to the target task. In our implementation, we fine-tuned the LLMs on 80\% of our film review corpus, using the remaining 20\% as a held-out test set to evaluate performance.

\begin{table}[h!]
\centering
\caption{Overview of methods tested and their variants.}
\label{tab:methods-variants}
\renewcommand{\arraystretch}{1.2}

\begin{tabular}{@{} L{3.2cm} L{3.6cm} C{3.1cm} L{5.0cm} @{}}
\toprule
\textbf{Method} & \textbf{Strategy} & \textbf{Number of examples} & \textbf{Creative vs conventional metaphor ratio} \\
\midrule

\multirow[t]{1}{*}{RAG} & \NA & \NA & \NA \\
\addlinespace[2pt]

\multirow[t]{9}{*}{Prompt engineering}
  & Zero-shot                     & 0 & \NA \\
  & \multirow[t]{4}{*}{Few-shot}  & 4 & Even \\
  &                                & 4 & Original \\
  &                                & 8 & Even \\
  &                                & 8 & Original \\
  & \multirow[t]{4}{*}{Chain-of-thought} & 4 & Even \\
  &                                       & 4 & Original \\
  &                                       & 8 & Even \\
  &                                       & 8 & Original \\
\addlinespace[2pt]

\multirow[t]{1}{*}{Fine-tuning} & \NA & \NA & \NA \\

\bottomrule
\end{tabular}
\end{table}

Table \ref{tab:methods-variants} provides an overview of the methods tested and their corresponding variants. We tested these approaches using the models shown in Table \ref{tab:llm-list}. We compare models of different sizes to assess whether size affects performance on the metaphor identification task. The size of the LLMs developed by OpenAI is not publicly available, but reliable estimates put them at several orders of magnitude larger than the relatively small open-source models we selected; GPT 4.1, for instance, is estimated to contain 1.8 trillion parameters \citep{gd.}. The `mini' and `nano' variants represent the medium- and small-sized versions of their respective base models (GPT 4.1 and o3). Smaller models have a lower energy and environmental impact and are therefore preferable to larger ones. In addition, the open-source models selected can all be run locally on modern consumer-grade hardware, affording users greater control and privacy. Evaluating the performance of open-source models is also important for widening access to this technology and improving the replicability of LLM-based research. We also include both reasoning and non-reasoning models in our evaluation, where reasoning refers to the model's capacity to perform multi-step logical inferences beyond simple pattern recognition.


\begin{table}[h!]
\centering
\caption{Large Language Models used in experiments.}
\label{tab:llm-list}
\begin{tabular}{@{} l l l l c @{}}
\toprule
\textbf{Model} & \textbf{Parameter size} & \textbf{Accessibility} & \textbf{Developer} & \textbf{Reasoning capabilities} \\
\midrule
Llama 3.1      & 8 billion      & Open source   & Meta     & No  \\
Llama 3.2      & 3 billion      & Open source   & Meta     & No  \\
Llama 3.2      & 1 billion      & Open source   & Meta     & No  \\
Deepseek R1    & 8 billion      & Open source   & DeepSeek & Yes \\
GPT 4.1        & Not disclosed  & Closed source & OpenAI   & No  \\
GPT 4.1 mini   & Not disclosed  & Closed source & OpenAI   & No  \\
GPT 4.1 nano   & Not disclosed  & Closed source & OpenAI   & No  \\
o3             & Not disclosed  & Closed source & OpenAI   & Yes \\
o3 mini        & Not disclosed  & Closed source & OpenAI   & Yes \\
o4 mini        & Not disclosed  & Closed source & OpenAI   & Yes \\
\bottomrule
\end{tabular}
\end{table}

The models were accessed via their Application Programming Interfaces (APIs) to enable batch execution of prompts. For each method and prompt variant tested, we conducted five experimental runs to assess the consistency of the output. The `temperature' parameter was left at its default setting. In the case of fine-tuning, each run used a different sample for both the training and test sets to ensure that results were not biased by the specific composition of the data. Fine-tuning was not performed with the reasoning models because, unlike standard models, they are trained using multi-stage procedures that combine supervised learning with reinforcement learning, which relies on expert human feedback to develop step-by-step reasoning capabilities. In contrast, non-reasoning models can be fine-tuned using supervised learning alone (i.e. training the model on labelled examples without additional reinforcement or human feedback steps), a capability that is fully supported through the standard fine-tuning APIs.

While, as discussed above, LLMs can be used without advanced programming skills, accessing them via their API does require some basic programming knowledge. To ensure that our workflow is as accessible and replicable as possible, we have created a set of annotated Jupyter Notebooks that include and explain the Python code used to run our experiments. These resources are publicly available via the following GitHub repository: \url{https://github.com/Weihang-Huang/MetaphorIdentification}. The repository also contains the manually annotated corpus, the uncoded corpus, the codebook used for annotation and RAG prompting, and the full set of complete prompts tested in this study.

\section{Results}
\label{sec:results}
This section presents the results of our experiments. We begin by reporting quantitative data on the LLMs' performance across the tested methods and prompting strategies. We then qualitatively analyse the output of the best-performing method to shed light on the types of inconsistencies observed between humans and LLMs, and to gain insights into how both LLM performance and the metaphor annotation protocol might be improved. All data and code for the quantitative analysis are available on our GitHub repository.

\subsection{Model performance}
\label{sec:performance}
We evaluated the models' performance against our ``gold standard'' hand-coded corpus using `precision', `recall', and `F1' measures. Precision quantifies how many of the identified metaphors correctly match our corpus, recall captures how many of the metaphors in the corpus the model successfully identified, and F1 combines both into a single harmonic score. These scores were calculated at the token level, meaning that each individual token was assessed independently as either coded or uncoded, rather than requiring an exact span match in the case of multi-word metaphorical expressions. This approach allowed us to capture cases where the model's output overlapped only partially with the human annotations and to award proportional credit for these partial matches. All examples used in few-shot prompts were kept in the corpus but excluded from the performance assessment.

\begin{figure}[h!]
    \centering
    \includegraphics[width=0.9\textwidth]{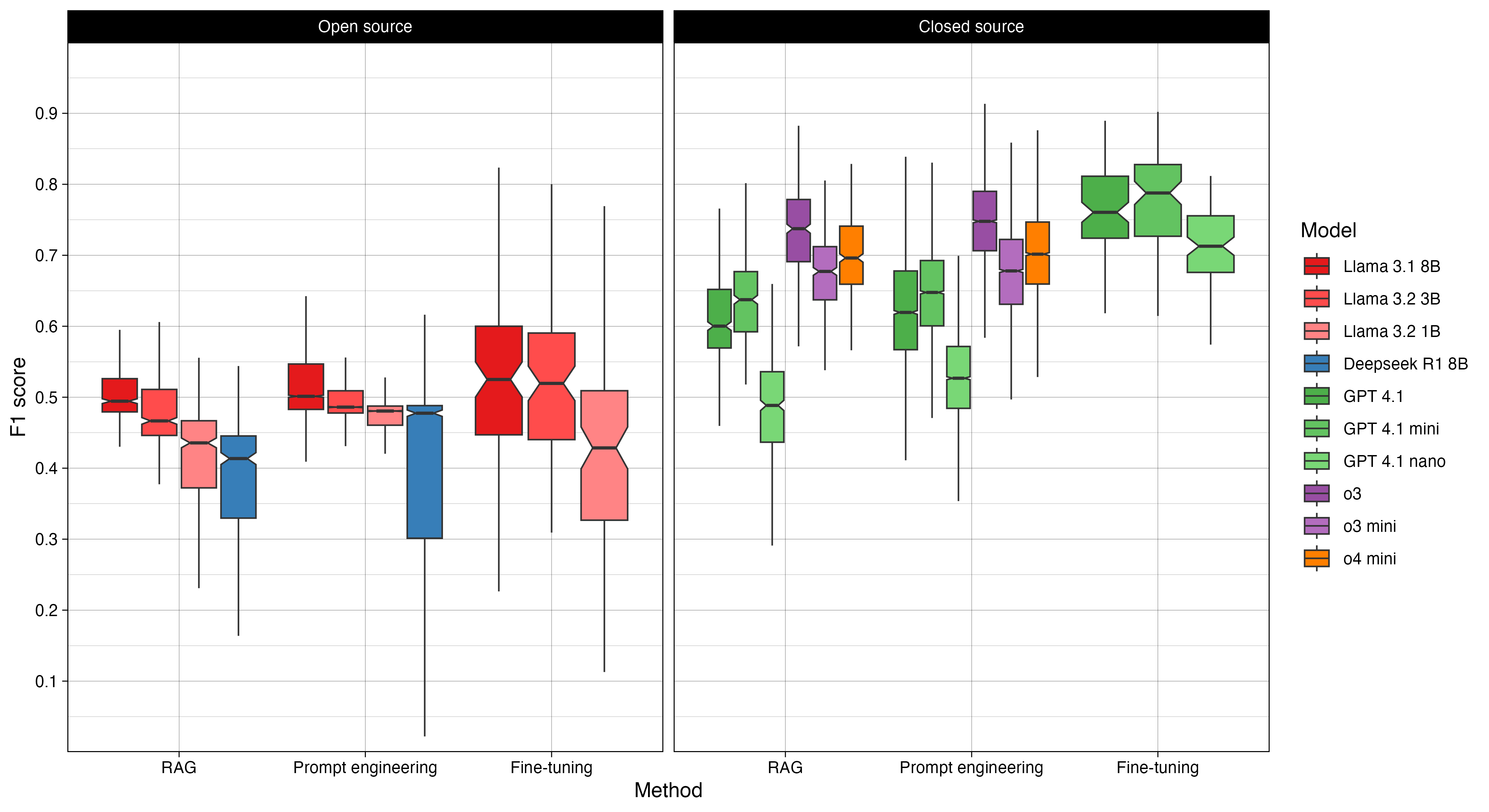}
    \caption{Median F1 scores with corresponding distributions across models and core methods. 
    The line inside each box represents the median, the box spans the interquartile range (IQR), 
    notches approximate 95\% confidence intervals for the median, and whiskers indicate 
    variability beyond the quartiles.}
    \label{fig:median-f1-methods}
\end{figure}

Figure \ref{fig:median-f1-methods} shows the median F1 scores and distributions of model performance across the three core methods and the 10 LLMs tested. Fine-tuning yielded the highest accuracy, followed by prompt engineering, and RAG. Within fine-tuning, the best-performing model was GPT-4.1 mini, which achieved a median F1 score of 0.79 (IQR: 0.73–0.83). The reasoning o3 model achieved the highest performance for both prompt engineering (median: 0.75, IQR: 0.71–0.79) and RAG (median: 0.74, IQR: 0.69–0.78). Overall, Figure \ref{fig:median-f1-methods} shows that closed-source models outperformed open-source models, with the latter displaying not only lower F1 scores but also greater variability in accuracy.

To test whether these differences are statistically significant, we fit a mixed-effects beta regression model using the \emph{glmmTMB} package in R \citep{bkvmbnsmb17}. The model included method, model type (open- vs closed-source), and text length as fixed effects, with random intercepts for model, text ID, and experimental run (i.e. the five repetitions of each approach). F1 scores were adjusted using the Smithson–Verkuilen transformation to fit within the (0,1) interval \citep{smithson2006better}, and text length was centred to improve model stability. 

The analysis revealed a significant effect of method on F1 scores. Compared to RAG, both prompt engineering ($\beta$ = 0.11, p \textless{} 0.001, $\Delta$F1 = 0.02) and fine-tuning ($\beta$ = 0.43, p \textless{} 0.001, $\Delta$F1 = 0.10) achieved significantly better performance, with fine-tuning also reliably outperforming prompt engineering ($\beta$ = 0.33, p \textless{} 0.001, $\Delta$F1 = 0.07). Overall, closed-source models performed substantially better than open-source ones ($\beta$ = 0.84, p \textless{} .001, $\Delta$F1 = 0.20). As shown in Figure \ref{fig:median-f1-methods}, closed-source models were also more responsive to the choice of method. For instance, GPT-4.1 showed much larger estimated gains when moving from RAG to fine-tuning ($\beta$ = 0.73, p \textless{} .001, $\Delta$F1 = 0.15) compared to an open-source model such as Llama 3.1 8B ($\beta$ = 0.08, p \textless{} .001, $\Delta$F1 = 0.02). Text length had a significant negative effect ($\beta$ = -0.10, p \textless{} 0.001, $\Delta$F1 = -0.02 per one SD increase), indicating that longer texts were more challenging to annotate accurately.

\begin{figure}[h!]
    \centering
    \includegraphics[width=0.9\textwidth]{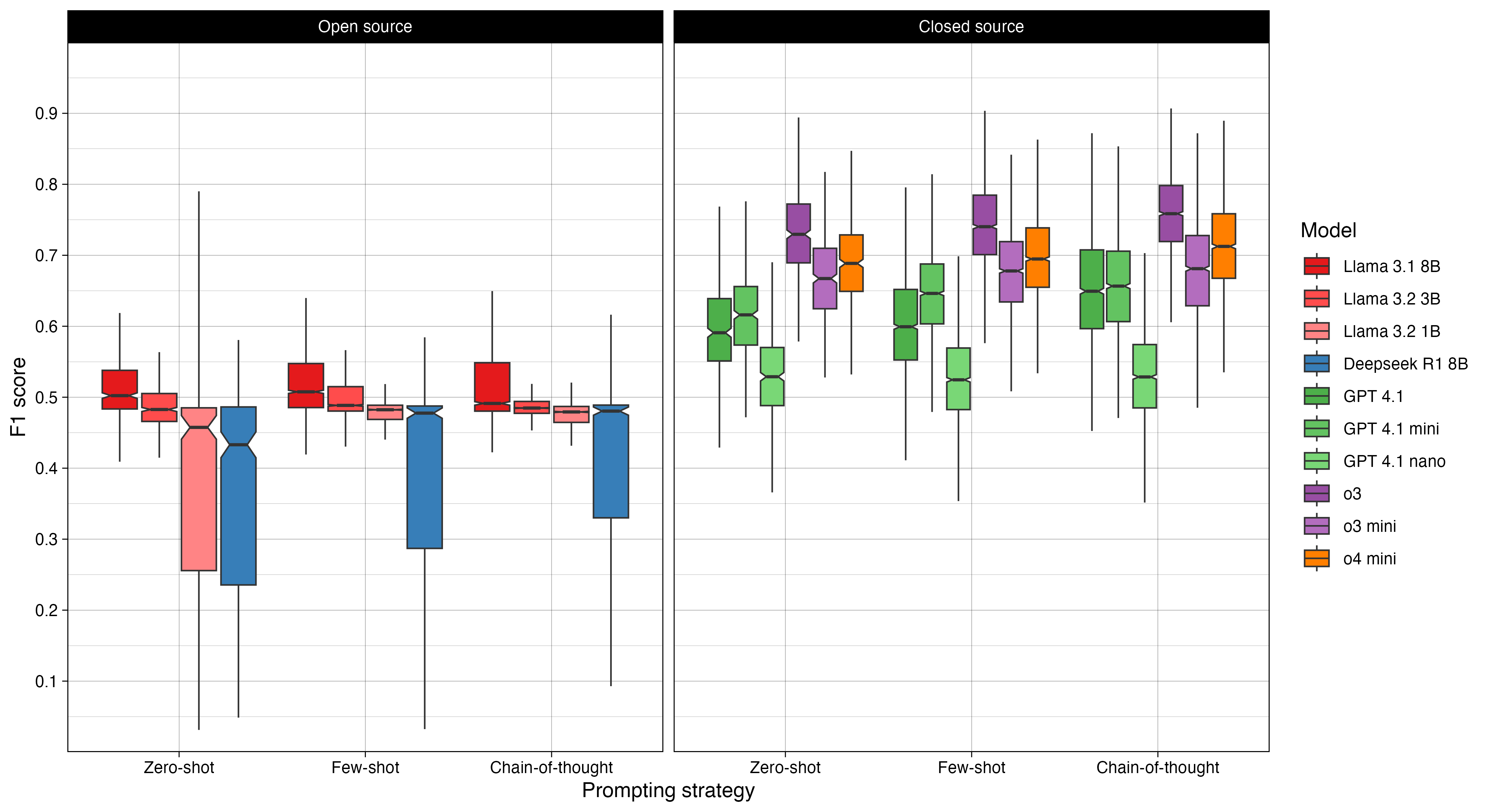}
    \caption{Median F1 scores with corresponding distributions across models and core prompt engineering strategies. The line inside each box represents the median, the box spans the interquartile range (IQR), notches approximate 95\% confidence intervals for the median, and whiskers indicate variability beyond the quartiles.}
    \label{fig:median-f1-prompts}
\end{figure}

Next, we zoomed in on prompt engineering to examine performance differences across the three core prompting strategies. Figure \ref{fig:median-f1-prompts} reveals a clear performance gap between open-source and closed-source models, with the former showing both lower accuracy and less sensitivity to the prompting strategy. Among the closed-source models, we observed a marked difference between reasoning and non-reasoning models, with the former substantially outperforming the latter. The performance of closed-source models progressively improved across the three prompting strategies, with chain-of-thought prompts consistently outperforming the others. This suggests that, at least for large closed-source models, providing examples and explicit reasoning steps does enhance performance. The best-performing configuration used the o3 model with chain-of-thought prompting, achieving a median F1 score of 0.76 (IQR: 0.72–0.80). These findings indicate that chain-of-thought prompting with best-in-class reasoning LLMs can achieve accuracy levels comparable to fine-tuning, but with the significant advantage of requiring only a small number of annotated examples. Even with a basic zero-shot prompt, closed-source reasoning LLMs demonstrated robust performance, with o3 achieving a median F1 score of 0.73 (IQR: 0.69–0.77). This indicates that these LLMs already have a rich internalized ``knowledge'' that enables them to carry out this task quite successfully.

To assess whether differences between the three core prompting strategies were statistically significant, we fitted a mixed-effects beta regression model with prompting strategy, model type, and text length as fixed effects, and random intercepts for model, text ID, and experimental run. As above, F1 scores were adjusted using the Smithson–Verkuilen transformation, and text length was mean-centred. The results confirm a significant effect of prompting strategy on model performance. Compared to the zero-shot baseline, both few-shot prompting ($\beta$ = 0.10, p \textless{} .001, $\Delta$F1 = 0.02) and chain-of-thought prompting ($\beta$ = 0.20, p \textless{} .001, $\Delta$F1 = 0.05) led to significant improvements in F1 scores. Chain-of-thought prompting yielded the best results, significantly outperforming not only zero-shot prompting but also few-shot prompting ($\beta$ = 0.10, p \textless{} .001, $\Delta$F1 = 0.02). This result suggests that adding both examples and explanations substantially enhances performance. Closed-source models significantly and substantially outperformed open-source ones ($\beta$ = 0.84, p \textless{} .001, $\Delta$F1 = 0.20). As observed above, text length was negatively associated with performance ($\beta$ = -0.15, p \textless{} .001, $\Delta$F1 = -0.04). 

Lastly, we tested the effects of (i) increasing the number of examples included in the prompt and (ii) varying the ratio of creative to conventional metaphors within those examples. The results show that both including four examples ($\beta$ = 0.13, p \textless{} .001, $\Delta$F1 = 0.03) and eight examples ($\beta$ = 0.16, p \textless{} .001, $\Delta$F1 = 0.04) led to significant improvements in model performance compared to the zero-shot baseline. There was also a statistically reliable but small improvement when increasing the number of examples from four to eight ($\beta$ = 0.03, p \textless{} .001, $\Delta$F1 = 0.01), indicating a marginal benefit from additional in-context data but with diminishing returns as the number of examples grows. Example sets that preserved the original ratio of conventional-to-creative metaphor ratio yielded a small but significant performance gain over those with an even ratio ($\beta$ = 0.04, p \textless{} .001, $\Delta$F1 = 0.01).

\subsection{Analysis of human-LLM discrepancies}
\label{sec:discrepancies}
After completing the quantitative evaluation of LLM performance, we turned to a more in-depth analysis of the discrepancies between the LLM-generated and human-generated metaphor coding. We focused on the annotations produced by the fine-tuned GPT 4.1 mini model, both because it achieved the highest accuracy and because its output is of a manageable size, as 80\% of the corpus was used to train the LLM. The sample we examined comprises 19 texts. The aim of this analysis was to identify the nature of the discrepancies, to better understand the possible reasons for them. 

We examine both `false negative' discrepancies, where the human coders identified a stretch of text as metaphor but the model did not, and `false positive' discrepancies, where the model identified a stretch of text as metaphor when the human coders did not. Two coders (both authors of this paper) worked together to identify reasons for each instance of a discrepancy. They began by discussing 10\% of the discrepancies in order to create a preliminary set of categories. They then coded the remaining discrepancies independently in Nvivo \citep{l25}, discussing and refining the categories as they did so. They then met for a number of joint coding sessions where they discussed and resolved borderline cases. The full list of possible reasons that were identified for false negative and false positive discrepancies is shown in Table \ref{tab:discrepancy-types}. Note that a single word or stretch of text could be coded into more than one category. As a result, the numbers reported here should not be interpreted as precise measures of output accuracy (which is instead provided above) but rather as heuristics for identifying the predominant sources of errors in the LLM coding. In a small number of cases (26), it was impossible to propose a reason for the discrepancy, so they were not included in the analysis below.

\begin{table}[h!]
\centering
\caption{Types of discrepancies observed in output from o3, fine-tuning method.}
\label{tab:discrepancy-types}
\begin{tabular}{@{} l rr @{}}
\toprule
\textbf{Discrepancy Type} & \textbf{False negative} & \textbf{False positive} \\
\midrule
Decomposability                                           &  5  &  4  \\
Degree of conventionality                                 & 59  & 12  \\
Difficulties in perceiving a source--target distinction   & 34  & 56  \\
Explicit comparisons                                      & 10  &  0  \\
Grammatical uses that cue metaphoricity                   & 21  &  7  \\
Difficulties in perceiving a relationship based on comparison &  0  & 12  \\
Personification                                           & 30  &  7  \\
Phrase-level meaning                                      & 22  & 26  \\
Source target confusion and twice true                    &  8  &  6  \\
\midrule
\textbf{Total}                                            & \textbf{189} & \textbf{129} \\
\bottomrule
\end{tabular}
\end{table}

Below we discuss each of these reasons in turn, beginning with those that were more likely to lead to false negative discrepancies, then discussing those that led, more or less equally, to discrepancies in both directions and ending with those that were more likely to lead to false positive discrepancies. In each sub-section, the reasons are discussed in order of their occurrences, beginning with those that occurred most frequently.

\subsubsection{Principal reasons for false negative discrepancies}
\paragraph{Degree of conventionality.}
A first reason for false negative discrepancies was that the model appeared to miss metaphors that are highly conventional in English, especially metaphors that do not have a literal alternative, such as those used for time. For example, in a number of cases, the model failed to identify cases of metaphors where time was expressed in terms of space, as in the case of \textit{the very \textbf{near} future} and cases of metaphor where time was expressed as a resource, as in: \textit{anyone who hasn't \textbf{spent} their life in seclusion}. Many of the instances identified in this category involved source domains relating to size, space, or containment (e.g. \textit{the film contains}\dots). 

\paragraph{Personification.}
A second reason for false negative discrepancies was that the model appeared to miss metaphor that could be considered to involve personification. An example of one such item in this category is: \textit{the fury mother nature can unleash}, where the whole phrase was missed by the model. Here the human coders saw \textit{nature} as personified as a (female) human capable of expressing emotions (in this case \textit{fury}).  

\paragraph{Explicit comparisons.}
A third source of discrepancy where the model missed examples that had been coded as metaphor by the human coders was that of explicit comparisons. In some cases, the metaphors were flagged in such a way as to make them resemble similes. For example, one of the films under review is described as being \textit{about as funny as a root canal}. This was coded as metaphorical by the human coders, as a metaphorical comparison is being drawn between the experience of watching the film and the experience of having painful dental treatment. In this case, the model may have been operating on the premise that similes and metaphors are different entities; a piece of folk wisdom that is often taught in schools. However, this distinction (between similes and metaphors) is somewhat arbitrary as it simply singles out two ways of signalling that a metaphorical comparison is intended (\textit{like} and \textit{as}) and treats them as if they are somehow ``special''. In reality, intended metaphorical meanings are signalled in many different ways, through the use of `tuning devices' such as \textit{kind of}, \textit{sort of}, \textit{metaphorically speaking}, and even the word \textit{literally} \citep[see][]{cd03}. For this reason, many metaphor researchers do not draw a tight distinction between metaphors and similes, although some coding schemes such as the MIPVU \citep{sdkkh10} draw a distinction between `indirect' metaphors (metaphors that are not signalled as such) and `direct' metaphors (metaphors that are signalled as such).

\paragraph{Grammatical uses that cue metaphoricity: crossing word class boundaries.}
The final source of false negative discrepancies related to examples where the grammatical structure of the language seemed to cue metaphoricity, a phenomenon that was first observed by \citet{deignan2005metaphor}. For example, the use of words of particular word classes at times favoured metaphorical interpretation. One example of this category is found in the expression \textit{things}\dots \textit{gave me \textbf{glassy} eyes} (i.e. they brought the writer to tears). This is a way of describing tear-filled eyes by likening them to a pane of glass. However, the adjectival derivation of \textit{glass} is always metaphorical, which may explain why the model missed this metaphor. 

Another example that relates to this category is the use of the word \textit{approach} in the sentence \textit{I found this \textbf{approach} to be very clever}. The nominal use of \textit{approach} tends to be associated with metaphorical meanings, and the verbal use with literal meanings. Again, the model did not code this example as metaphorical, although the human coders did. 

This category was also used for cases where the metaphoricity of the term came out of its grammatical role in the sentence. For example, the phrase \textit{bathed in}, when used in a literal context (e.g. \textit{I bathed in the sea}) tends to carry a human actor. However, when used metaphorically (e.g. \textit{she was bathed in soft light}), the phrase tends to appear in the passive voice. The model tended to miss such cases, suggesting that human coders were more attuned to grammatical cues that signal metaphoricity.

\subsubsection{Principal reasons for discrepancies that occurred in both directions}
In some cases, the discrepancies did not skew towards one type or another. There were four main reasons behind such discrepancies.

\paragraph{Difficulties in perceiving a source-target distinction.}
The first reason for discrepancies seen equally in false negative and false positive contexts occurred when it was difficult to draw a distinction between the source and target domains. These examples lie on the cusp of being literal or metaphorical. For example, in the following extract, the model coded the phrase \textit{join the list} as metaphorical:

\begin{quote}
With many big-budget science fiction films, great ideas are often wasted by bad scripts, cheesy plot twists, and terrible acting. The Fifth Element, The Abyss, and Godzilla had great concepts squandered by bad acting, writing, or both. At first glance, The Matrix, Larry \& Andy Wachowski's sci-fi/kung-fu/shoot-em-up spectacular, looks like a prime candidate to \textbf{join the list} of high-concept bad movies [\dots]    
\end{quote}

In this example, one might say that the reviewers had actually produced a list of three \textit{high concept bad movies} so the use of the term list here is, in some ways, literal. On the other hand, there is a degree of ambiguity in this example. The so-called list may just have been a way of saying that these three movies were all terrible. When the review talks about the list of high concept bad movies, there is a degree or irony and a faint suggestion that this is some sort of ``official'' list, which of course does not exist. The model may have picked up on this idea, which goes some way towards explaining why it coded it as metaphorical: because an ``official'' list such as this does not actually exist.

In a second example, the model coded \textit{prediction game} as metaphorical in the following sentence:

\begin{quote}
You can tell a script is terrible when you are able to predict what will happen minutes before it does. This little \textbf{prediction game} is a very fun exception.
\end{quote}

Here the viewer may well be playing a game with themselves; although the game itself does not meet all the criteria for being a prototypical game, it does meet some of them, as the activity is fun and playful, and involve a degree of chance. 

A number of items in this category involved the use of sight terms as a source domain (e.g. \textit{examining} concepts). As \citet{l15} discuss, there is often a degree of conflation between the literal and the metaphorical in cases such as these. Interestingly, this conflation seems to have been picked up by our coding scheme. Sometimes, a potential relationship of metonymy could be perceived in these examples. For example, in the phrase \textit{\textbf{pounding} drum beats}, there is no clear separation between the source and target domains; instead, the action of \textit{pounding} a drum is used to describe the resulting sound.

\paragraph{Phrase level meaning.}
A second reason for the discrepancies that occurred in both directions was related to the fact that it is not always clear whether the metaphoricity lies at the level of the word or at the level of the phrase. In such cases, both the model and the human coders sometimes tagged the whole phrase as metaphorical, and sometimes simply tagged the part of the phrase that was most saliently metaphorical. For example for the phrase \textit{a member of that undistinguished club}, the model identified the word \textit{club} as metaphorical whereas the human coders coded the whole phrase as such. Similarly, in the phrase, \textit{left behind in the romantic wake}, the model identified \textit{left behind} and \textit{wake}, but not the other words in the phrase. The human raters coded the whole phrase as metaphorical as it represented a single metaphorical scene.

\paragraph{Decomposability.}
A third reason for the discrepancies that occurred in both directions was that in some cases the metaphorical component could be found only in one part of a lexical item and could therefore be considered to operate below the level of the word. An example of this is the term \textit{eye-catching}, which was identified as metaphorical by the model but not by the human coders. Here the word \textit{eye} is used metonymically to refer to the act of seeing but the term \textit{catching} is more metaphorical as it reifies the reception of the visual information.

\paragraph{Source-target confusion and twice-true metaphors.}
A fourth reason for discrepancies that appeared in both directions concerned metaphorical expressions that were also literally true in the context of the film reviews. This situation proved challenging for both the human coders and for the model. This is clearly a grey area in metaphor identification, and perhaps needs to be addressed explicitly when instructing LLMs in the practice of metaphor identification. One example of a word that was coded by the human coders as metaphorical but not the model was the word \textit{dark} in the following passage:

\begin{quote}
While the original Alien film was a dark, enclosed horror film [\dots]
\end{quote}

Here, the word \textit{dark} could be being used literally but it could also be being used metaphorically, and in fact was probably being used in both ways.

Another example of a stretch of text that was coded as metaphorical by the human coders but not by the model involved the phrase \textit{tells the tale} in the following sentence:

\begin{quote}
[\dots] The Mod Squad tells the tale of three reformed criminals under the employ of the police to go undercover.
\end{quote}

This sentence could be paraphrased as saying that the film [The Mod Squad] is \textit{about} three reformed criminals. However, the writer chooses to use the formulation \textit{tells the tale} instead. Although it is highly conventional to talk about a film telling a particular story, the idea of ``telling a tale'' creates an image of someone (perhaps a grandparent) perhaps sitting down in an armchair and starting to recount a long and traditional story. There is therefore potential for a metaphorical comparison here though it is by no means clearcut.

\subsubsection{Principal reasons for false positive discrepancies}
\paragraph{Difficulties in perceiving a contemporary relationship between the source and target that was based on comparison.}
There was only one clear reason for false positive discrepancies. This related to the ease with which it was possible to identify a contemporary relationship between the source and target that was clearly based on similarity and comparison. For example, the model coded \textit{red herring} as metaphor whilst the human coders did not do so. Whilst the expression \textit{red herring} is clearly not literal, it is difficult to understand its meaning in context by identifying any kind of similarity between misleading clues and red herrings. This may be due to the metaphors being historically older, where the historical mappings have fallen away.

\subsubsection{Discussion of human-LLM discrepancies}
Interestingly, many of the discrepancies closely mirror issues that are often discussed in the metaphor literature. The first issue concerns the nature of the different types of figurative language, and the relationships between them. We saw examples above of discrepancies relating to the blurred relationship between personification (\textit{the fury mother nature can unleash}) and metaphor, and between metonymy and metaphor (\textit{eye-catching}). There is some debate over whether personification should be viewed as a kind of metaphor, and a persuasive body of literature showing how personification can cross the spectrum from metaphorical to literal language \citep{d11}. 

The second issue concerns the level at which metaphor operates. In the metaphor literature, there is much debate over whether metaphor operates at the level of the word or at the level of the phrase, and over the role played by metaphorically used words within a whole phrase. Some words contribute disproportionately to the overall metaphoricity of a phrase, and the model appears to be picking these words up. However, the human coders had a stronger tendency to code the whole phrase as metaphorical as it was thought to constitute a single metaphorical scene \citep{jo23}. Equally, most metaphor coding schemes do not allow for cases where metaphor is operating below the level of the lexical unit, and this includes cases of hyphenated words, such as \textit{eye-catching}.

Conventionality sometimes shades into semantic bleaching. Some words do not retain sufficiently strong links to their original concrete meaning to be considered metaphor. For instance, metaphors representing time in terms of space fall into this category as they are often highly conventional and therefore have the potential to be easily be missed by both human coders and models. Issues such as these, which are pervasive throughout the metaphor literature, explained a large number of the inconsistencies between the human and the LLM coding. In both types of coding, ``rules'' relating to the relationship between metaphor and other types of figurative language, the level at which metaphor could be located (word, phrase, or sub-lexical unit), and conventionality were difficult to apply as all of these phenomena operate along continua and do not fall neatly into discrete categories. Humans struggle with artificial dichotomies, and perhaps because LLMs are based on human data, they also appear to struggle with them. This finding is unsurprising. The very act of attempting to say whether or not a particular language item is or is not metaphorical is an artificial endeavour as language does not operate in either/or categories. Metaphoricity, like any other aspect of language is best seen as a radial category with more or less prototypical examples \citep{t03} and there are no universal, context-independent cut-off points. Therefore, the fact that LLMs fail to consistently apply false dichotomies renders them human-like.

\section{Conclusion}
\label{sec:conclusion}
As a pervasive feature of discourse, metaphor offers a productive lens for examining cognition, emotion, and ideology. Yet large-scale analysis has so far been limited by the need for manual annotation, which stems from the context-sensitive nature of metaphor. In this study, we explored the potential of LLMs to automate metaphor identification, with the ultimate aim of making metaphor analysis more scalable and its findings more robust and generalisable. We assessed, for the first time, the ability of LLMs to detect and annotate metaphors in full texts. This approach offers greater ecological validity than previous work, which largely relied on single, decontextualized sentences, and produces fully annotated texts that can be used directly in downstream analyses. In addition, we extended prior research by evaluating not only prompt engineering but also retrieval-augmented generation (RAG) and fine-tuning.

The results of our experiments demonstrate that state-of-the-art closed-source LLMs can identify and annotate metaphor in full texts with a high level of accuracy. The best median F1 score of 0.79, achieved through fine-tuning, is substantially higher than results from previous studies that evaluated LLMs on the simpler task of annotating individual words in isolated sentences \citep{liang2025gpt4, hicke2024science}, and comes close to the best scores reported for the even simpler task of annotating pre-selected single words \citep{prr24, txm24}. Crucially, humans themselves often disagree on what constitutes a metaphor. Even well-established metaphor identification protocols fall short of perfect inter-coder agreement, with reported kappa scores ranging from 0.56 to 0.70 for MIP \citep{g07} and from 0.74 to 0.88 for MIPVU \citep{sdkkh10}. Thus, while F1 and `kappa' are not directly comparable, it seems reasonable to conclude that LLM performance is not far from human levels of reliability. Our comparative analysis of human and LLM outputs lends further support to this conclusion by showing that most discrepancies are not random but reflect well-known grey areas and ontological challenges that metaphor scholars have long grappled with.

Based on these results, we propose that LLMs offer a viable means of at least semi-automating metaphor identification and annotation. As noted at the outset, LLMs provide an accessible and flexible way to efficiently annotate large text corpora, thereby freeing up time and resources for higher-level interpretation and theory-building. Crucially, we are not advocating the use of LLMs as ``black boxes'' whose outputs are accepted uncritically. Instead, we envisage a recursive, human-in-the-loop workflow consisting of three main stages. In the development stage, different LLM methods are tested and iteratively refined through close analysis of small output samples, building on best practices from previous research, including our own study. Once the most effective approach has been identified and its accuracy optimized, it can then be applied to the full corpus. Finally, if preliminary tests indicate that the model's accuracy falls short of the project's requirements, a human analyst can manually correct the LLM-generated annotations. Even when this step is necessary, it should be considerably less time-consuming and cognitively demanding than annotating the texts entirely from scratch. The manually corrected dataset could then be used to fine-tune an LLM to further expand the analysis. Given the robust results we obtained using a small corpus, we expect there to be considerable scope for further improvement with larger training datasets.

Crucially, however, we do not view LLMs merely as a methodological tool for increasing the scale of analyses. Our examination of discrepancies between LLM and human output highlights their potential as a testbed for developing and refining theoretical understandings of metaphor, including the boundaries between metaphor and other figurative tropes. In addition, LLMs can serve as a means to benchmark different metaphor annotation procedures and iteratively refine them, in much the same way that inter-coder reliability testing is used to evaluate and refine human annotation protocols \citep{f18}.

More broadly, our study contributes to the rapidly growing body of evidence that establishes LLMs as a useful tool for automating a wide range of linguistic annotation tasks \citep[e.g.][]{gf25, mn25, mm25, ylsf24, ybh24}. Our results offer a counterpoint to some of the concerns raised by \citet[p. 9]{cmb25} regarding the suitability of LLMs for corpus annotation. First, contrary to the authors' claim, our study shows that LLMs do not require extensive training data to achieve good performance (beyond, of course, the data used for pre-training the model itself). The hand-coded dataset used for fine-tuning in our study included only a few thousand words, and even with as few as eight annotated sentences we obtained high-quality results. Second, LLMs do not demand an excessive upfront time investment – certainly no more than the effort typically required to develop a codebook for manual annotation and to test its reliability through inter-coder agreement. Moreover, the LLM-based approach offers clear incremental advantages: now that we have carried out the groundwork of setting up and evaluating different methods and prompt variants, the most effective approach can be further refined in future iterations. Third, while LLM output does contain errors (just as human coding does), these errors are largely systematic and interpretable, making them amenable to targeted correction and improvement. Finally, while \citet{cmb25} are certainly right to highlight the environmental concerns associated with LLMs, these concerns should be weighed against the human costs and ethical implications of employing research staff – often graduate students and junior researchers – to carry out labour-intensive and tedious annotation tasks for extended periods of time.

Our study provides several promising directions for future research to further explore the potential and applications of LLMs in metaphor analysis. First, future work could build on our analysis of discrepancies between human and LLM output by refining prompts and providing tailored instructions and examples to reduce ambiguity and improve accuracy. The three main methods we tested could also be combined to assess whether such integration has a multiplier effect on performance. Second, future studies could replicate our experiments using alternative annotation procedures, such as MIP(VU), to determine which performs best and to test whether our improved results compared to previous LLM-based studies are due to our techniques and prompts or to the phraseological approach we adopted. Put differently, is a phraseological approach to metaphor inherently more intuitive, and does this explain why we obtained better results than earlier studies relying on MIP(VU)? Third, the approaches developed here could be applied and evaluated in other discursive domains to examine whether factors such as genre or topic influence LLM performance. Additional LLMs could also be tested, including large open-source models such as Llama 3 405B and Mistral 124B. Finally, while our study focused on metaphor identification, an equally important and challenging task for future LLM evaluation is metaphor labelling: identifying the metaphorical target and source domains. In sum, our findings suggest that LLMs have the potential to become a valuable component of the metaphor analysis workflow, and we hope this study encourages further research into this emerging area.

\section*{Acknowledgments}
We are grateful to Jack Grieve and Marcus Perlman (University of Birmingham) for their helpful feedback and discussions on this study.

\bibliographystyle{apalike}
\bibliography{references}  






\end{document}